\documentclass[sigconf, screen, a4paper]{acmart}
\usepackage{amsmath}
\usepackage{bbm}
\usepackage{xcolor}         
\usepackage{mathtools}
\usepackage{siunitx}
\usepackage{dblfloatfix}
\usepackage{graphicx} 
\graphicspath{{figs/}}
\usepackage{pifont}
\usepackage[inline,shortlabels]{enumitem} 
\usepackage{subcaption}
\usepackage{marvosym}
\usepackage{etoolbox}
\makeatletter

   \vspace{5pt}

\makeatother
\DeclarePairedDelimiterX{\infdivx}[2]{\Big(}{\Big)}{%
  #1\;\delimsize\|\;#2%
}
\newcommand{\infdiv}{KL\infdivx}

\renewcommand{\eqref}[1]{Eq.\,(\ref{#1})}

\DeclareMathOperator{\Prob}{\mathrm{P}}
\DeclareMathOperator{\enc}{Enc}
\DeclareMathOperator{\dec}{Dec}

\renewcommand{\S}[1]{\mathcal{S}_{#1}}
\newcommand{\Z}{\mathcal{Z}}
\renewcommand{\vec}[1]{\mathbf{#1}}
\newcommand{\x}[1]{\vec{x}_{t_{#1}}}
\newcommand{\xrec}[1]{\tilde{\vec{x}}_{#1}^{\S{1}}}
\newcommand{\z}[1]{\vec{z}^{\S{#1}}}

\newcommand{\T}{^{\top}}
\newcommand{\TS}[1]{^{\S{#1}}}

\newcommand{\Tset}[1]{\mathcal{T}_{#1}}

\newcommand{\yA}[1]{\hat{y}_{\text{obs}}}
\newcommand{\yR}[1]{\hat{y}_{\text{rec}}}

\newcommand{\romup}[1]{\uppercase\expandafter{\romannumeral #1\relax}}
\newcommand{\romlo}[1]{\lowercase\expandafter{\romannumeral #1\relax}}
\newcommand*{\argIn}{\makebox[1ex]{\textbf{$\cdot$}}}

\newcommand{\Rset}[1]{\mathbb{R}^{#1}}
\newcommand{\ts}{\textsuperscript}

\newcommand{\ie}{\textit{i}.\textit{e}.,}
\newcommand{\eg}{\textit{e}.\textit{g}.,}
\newcommand{\cf}{\textit{cf}.}

\newcommand{\rom}[1]{\uppercase\expandafter{\romannumeral #1\relax}}
\newcommand{\romsm}[1]{\lowercase\expandafter{\romannumeral #1\relax}}
\newcommand{\iid}{\textit{i}.\textit{i}.\textit{d}.}

\newcommand{\blue}[1]{{#1}}
\newcommand{\myAlgorithm}{\underline{O}nline \underline{L}earning \underline{D}eep models from \underline{D}ata of \underline{D}ouble \underline{S}treams}
\newcommand{\myAlg}{OLD\ts{3}S}
\copyrightyear{2022}
\acmYear{2022}
\setcopyright{rightsretained}
\AtBeginDocument{%
 \providecommand\BibTeX{{%
   \normalfont B\kern-0.5em{\scshape i\kern-0.25em b}\kern-0.8em\TeX}}}

\acmSubmissionID{2752}

\copyrightyear{2022}
\acmYear{2022}
\setcopyright{rightsretained}
\acmConference[MM '22]{Proceedings of the 30th ACM International Conference on Multimedia}{October 10--14, 2022}{Lisboa, Portugal}
\acmBooktitle{Proceedings of the 30th ACM International Conference on Multimedia (MM '22), Oct. 10--14, 2022, Lisboa, Portugal}
\acmISBN{978-1-4503-9203-7/22/10}
\acmDOI{10.1145/3503161.3548355}
\begin{document}

\title{Online Deep Learning from Doubly-Streaming Data}

\author{Heng Lian}
\email{hlian001@odu.edu}
\affiliation{%
  \institution{Old Dominion University}
  \streetaddress{5115 Hampton Boulevard}
  \city{Norfolk}
  \state{Virginia}
  \country{USA}
  \postcode{23529}
}

\author{John Scovil Atwood}
\email{jatwo002@odu.edu}
\affiliation{%
  \institution{Old Dominion University}
  \streetaddress{5115 Hampton Boulevard}
  \city{Norfolk}
  \state{Virginia}
  \country{USA}
  \postcode{23529}
}

\author{Bo-Jian Hou}
\email{houbo@pennmedicine.upenn.edu}
\affiliation{%
  \institution{University of Pennsylvania}
  \city{Philadelphia}
  \state{Pennsylvania}
  \country{USA}
}

\author{Jian Wu}
\email{j1wu@odu.edu}

\affiliation{%
  \institution{Old Dominion University}
  \streetaddress{5115 Hampton Boulevard}
  \city{Norfolk}
  \state{Virginia}
  \country{USA}
  \postcode{23529}
}

\author{Yi He}
\email{yihe@cs.odu.edu}
\authornote{Corresponding author: Dr. Yi He (yihe@cs.odu.edu)}
\affiliation{%
  \institution{Old Dominion University}
  \streetaddress{5115 Hampton Boulevard}
  \city{Norfolk}
  \state{Virginia}
  \country{USA}
  \postcode{23529}
}

\renewcommand{\shortauthors}{Heng Lian et al.}
\begin{abstract}
    This paper investigates a new online learning problem
    with doubly-streaming data,
    where the data streams 
    are described by feature spaces that constantly evolve,
    with new features emerging and old features fading away.
    %
    %
    A plausible idea to deal with such data streams
    is to establish a relationship
    between the old and new feature spaces,
    so that an online learner can leverage 
    the knowledge learned from the old features
    to better the learning performance on the new features.
    Unfortunately, this idea does not scale up to 
    high-dimensional multimedia data 
    with complex feature interplay,
    which suffers a tradeoff between {\em onlineness},
    which biases shallow learners,
    and {\em expressiveness}, which requires deep models.
    %
    %
    %
    Motivated by this,
    we propose a novel \myAlg\ paradigm,
    where a shared latent subspace is discovered 
    to summarize information from the old and new feature spaces,
    building an intermediate feature mapping relationship.
    %
    A key trait of \myAlg\ is to treat
    the {\em model capacity} as a learnable semantics,
    aiming to yield optimal model depth and parameters {\em jointly}
    in accordance with the complexity and non-linearity of the input
    data streams in an online fashion.
    Empirical studies
    substantiate the viability and effectiveness of our proposed approach.
    The code is available online at \url{https://github.com/X1aoLian/OLD3S}.
\end{abstract}



\maketitle

\section{Introduction}

Machine learning has become a fundamental building block 
in many cyber infrastructures,
provides an automated hence scalable apparatus 
to analyze the high-dimensional data streams 
(\eg\ images, texts, videos)
pervading all corners of the Internet~\cite{gong2007machine,jiang2016web,gao2017learning}.
Examples include
multimedia retrieval~\cite{tousch2012semantic,thomee2012interactive},
online speech analytics~\cite{fortuna2018survey,ferrara2019history},
recommender systems~\cite{chen2017attentive,wu2019deep,wu2019posterior,wu2020data,deldjoo2020recommender},
to just name a few.
Generally speaking, wherever it is infeasible
to inspect and process the data growing in an 
increasingly unmanageable volume with manpower,
machine learning prevails.

Despite their fashionability,
a prominent drawback shared by most existing machine learning methods 
is their limited {\em generalization capability}~\cite{marcus2020next}. 
As a matter of fact, machine learning models usually do well in practice
only if the data arriving in future tend to follow a nearly identical 
distribution as the data they were trained on~\cite{bousquet2002stability,kawaguchi2017generalization}.
This so-called \iid\ assumption inevitably limits the 
model expressiveness to our society that constantly evolves.

To aid the situation,
a new learning paradigm 
termed \emph{online learning from doubly-streaming data} 
has emerged with both %
algorithmic designs~\cite{hou2017learning,hou2017one,beyazit2019online,he2019online,zhang2020learning,hou2020storage,9406178,he2020toward,he2021online,he2021unsupervised,he2021onlineICDM} 
and domain applications~\cite{zhang2016inferring,xiao2017multiple,cano2020kappa,li2019multistream,nie2020online}.
Its key idea is 
to generalize learning models in two spaces.
First, the {\em sample space}, where the data instances are 
generated ceaselessly,
requiring to train learners on-the-fly, 
making real-time predictions as the data arrive.
As such, if the patterns underlying data changed,
an online learner can be updated instantly 
to adapt to the shift,
thereby retaining its accuracy performance over time~\cite{gama2014survey,lu2018learning,zhang2020online}.

Second, the {\em feature space}, 
where sets of features describing the arriving data samples
evolve, with new features emerge and old features 
stop to be generated.
To wit, a smart manufacturing pipeline
may employ a set of sensing techniques
to detect unqualified products~\cite{huang2021grassnet},
where each sensor coheres to a feature.
The feature space evolves,
when the old sensors wear out
and a batch of new sensors are deployed~\cite{hou2017learning}.
Tangibly, as the new and old sensors (\ie\ features) 
often differ in terms of amount, version, metric, and positions,
a new classifier needs to be initialized.
Yet, this new classifier may stay \emph{weak} and error-prone
before the training samples carrying these new features
grows to a sufficiently large volume.
Meanwhile, the old classifier becomes unusable
with the unobserved features,
leading to substantial waste of 
the data collection and training effort.
%
%
%
%
A 
relationship between the pre-and-post evolving feature spaces
must be established,
so that the old features can be {\em reconstructed} from the new ones.
Online learners can thus harvest the information 
embedded in the old classifier to aid the weak new classifier,
enjoying a boosted learning performance~\cite{hou2020storage,he2020toward,he2021online,he2021onlineICDM}.

Unfortunately, all existing studies 
suffer from a tradeoff 
between \emph{onlineness} and \emph{expressiveness}.
Specifically, on the one hand,
shallow learners (\eg\ generalized linear models~\cite{zinkevich2003online}, Hoeffding trees~\cite{schreckenberger2020restructuring})
possess a faster online convergence rate,
thanks to their simple model structures with a
small number of trainable parameters~\cite{shalev2012online}.
However, due to their limited learning capacity,
they usually end up with inferior performance when dealing with
high-dimensional media streams,
of which the feature interplay is often complex.

On the other hand, 
deep learners 
(\eg\ neural networks~\cite{lecun2015deep,ijcai2018-369}, deep forests~\cite{zhou2019deep,ren2019deep})
enjoy a low-dimensional hidden representation to
build accurate predictive models on complex raw inputs.
Yet, their large number of parameters 
residing in the entangled model structures
invites stochastic updates,
leading to a very slow convergence rate.
In an online learning context, more error predictions tend 
to be made before the learners converge
to an equilibrium.
%
%
These additional errors are recognized as \emph{regrets},
where the slower the convergence rate,
the larger the learner regrets in a hindsight.
%


Motivated by this tradeoff,
this paper mainly explores one question:
{\em 
How can we build an online learner that joins the two merits, namely,
1) converges as fast as shallow models
to minimize the online regrets and 
2) learns latent representations as expressive as deep models
from high-dimensional inputs with complex feature relationships.
}

Our affirmative answer provides a novel learning paradigm, termed 
\emph{\myAlgorithm} (\myAlg).
Our key idea is to train an online learner 
that \emph{automatically} adjusts its learning capacity
in accordance with the complexities 
and temporal variation patterns 
of input data stream.
Specifically, 
\myAlg\ is with an over-complete neural architecture~\cite{lewicki2000learning,van2021hedge,ijcai2018-369}
and starts from using its shallow layers,
approximating a simple classifier
to attain fast convergence at initial rounds.
Over time, the deeper layers are gradually mobilized, 
as more samples streaming in 
requires 1) a highly capable classifier that 
can learn expressive latent representations and 2) 
a precise delineation of complex feature interplay.
Knowledge reuse is enabled in 
both \romsm{1}) the shallow-to-deep model switch
via representations sharing 
and \romsm{2}) 
the pre-and-post evolving feature spaces
via reconstructive mapping and ensemble learning~\cite{zhou2021ensemble}.
This benefits our approach 
by expediting the convergence in a temporal continuum,
so as to maximize its online efficiency and efficacy 
when learning from doubly-streaming data.
{\bf Specific contributions} of this paper are summarized as follows:
\vspace{-1em}

\begin{enumerate}[label=\roman*)]

	\item This is the first study to explore 
	the doubly-streaming data mining 
	problem in an online deep learning context,
	where the high-dimensional data streams with feature space evolution 
	tend to incur a tradeoff between convergence rate and learning capacity.
	The technical challenges are manifested from empirical evidence in Section~\ref{sec:preliminary}.

	\item A novel \myAlg\ approach is proposed to tackle 
	the problem, where a modeling architecture
	with its depth \emph{learned} from data is devised
	to adapt to minimize the online classification regrets and 
	precisely approximate the feature-wise relationship
	on-the-fly. 
	%
	Details are in Section~\ref{sec:method}.
	
	%
	%
	

	\item Real-world high-dimensional datasets
	covering domains of machine translation and image classification
	are employed to benchmark our approach.
	Results suggest the viability and effectiveness of our proposal,
	documented in Section~\ref{sec:experiments}.
	
	
	\end{enumerate}

\section{Related Work}


%

\subsubsection*{\bfseries Online Learning with Doubly-Streaming Data}
%

Online learning algorithms were devised for
data stream processing~\cite{aggarwal2007data,shalev2011online},
where the reality of learning is in an on-the-fly setting
hence lifts the memory constraint for data analysis at scale.
In addition to allowing data to grow in terms of \emph{volume},
in an orthogonal setting,
hoping the \emph{features} describing input data to stay strictly unchanging 
is unrealistic over long time spans.
As a response,
the pioneering studies~\cite{zhang2015towards,zhang2016online,hou2018safe,beyazit2018learning,wu2021latent} 
explored a setting of {\em incremental} feature learning,
allow the arriving data instances to carry different sets of features 
yet later instances are assumed to include monotonically more features 
than the earlier ones. 
Subsequent works that strive to learn \emph{evolving} feature spaces~\cite{hou2017learning,hou2017one,beyazit2019online,he2019online,zhang2020learning,hou2020storage,9406178,he2020toward,he2021online}  
further relaxed the monotonicity constraint on the feature dynamics, 
enable effective learning when later instances stop carrying
old features that appeared theretofore.
A key technique shared by these methods
is to establish a mapping relationship between the old and new feature spaces.
As such, once the old features fade away,
their information can be reconstructed via the mapping,
aiding the weak learner trained on insufficiently few instances 
carrying new features,
join to make highly accurate predictions.

Despite their effectiveness in various settings,
these methods all prescribe a \emph{linear} model to fit the mapping,
which is unfortunately not capable to deal with complex real data,
\eg\ images in an evolving spectrum domain,
documents written in different languages.
We are aware of a very recent work~\cite{he2021onlineICDM} 
that does not use linear but copula model
to fit a non-linear mapping with statistical guarantees.
However, this work requires to deem each feature as a copula component,
and hence cannot scale up to a high-dimensional space 
(\eg\ images or natural languages).
Our proposed \myAlg\ approach does not suffer this restriction 
by discovering a latent feature space in which the original data dimension 
is largely condensed,
thereby being generalizable to a wider range of real applications.

\subsubsection*{\bfseries Deep Learning with Adaptive Capacity}

%
Neural networks have emerged for several decades to approximate
underlying functions with arbitrary complexity~\cite{chen1995universal,lu2020universal,lu2021learning}.
However, their universal approximation capability is grounded 
on an assumption of an infinitely wide hidden layer,
which cannot be satisfied in practical modeling.
The advent of Deep Neural Networks sidestepped this issue 
by imposing a \emph{hierarchical} representation learning procedure~\cite{bengio2013representation,lin2014stable,mhaskar2017and},
trading in width for depth,
so as to fit complex decision functions underlying data.
However, this hierarchical design introduces \emph{over-parameterization},
where the large number of learnable parameters request massive rounds of 
training iterations over huge datasets to converge.
Online decision-making using deep learning thus becomes seemingly impossible.

A key question to solve the challenge is how to choose the network depth
(representing the entire model capacity) in accordance with 
the underlying function in an adaptive, automated, and data-agnostic fashion.
Huang et al.~\cite{huang2016deep} firstly theorized and implemented the concept of stochastic depth, a training procedure that trains shallow networks and tests with deep networks, randomly dropping a subset of layers to quickly identify key layers. A method of deducing which layers can be trimmed is therefore needed.
Larsson et al.~\cite{larsson2016fractalnet} later identified a strategy to construct deep networks structured as fractals. This confers the ability to regularize co-adaptation of subpaths, effectively allowing for the isolation of high performing layers within a larger architecture. We can now judge values of groups of layers, making a delineation of value more concrete. 
Sahoo et al.~\cite{ijcai2018-369} and He et al.~\cite{he2021unsupervised}
demonstrated a {\em Hedge Backpropagation} mechanism for 
online/lifelong deep learning, 
where the model depth is deemed as a trainable 
semantic metric, jointly with the layer parameters to decide the  
function complexity learned from data streams in a dynamic way.


Unfortunately, all these deep methods fail to take 
the feature space evolution into account,
a factor that can largely affect the non-linearity of the
resultant learning function.
As a result, they cannot be adapted to learn 
the doubly-streaming data.
To fill the gap,
we propose to bring together the two fragmented 
subfields of online deep learning and doubly-streaming data mining.
In particular, we respect that the mapping relationship between 
the pre-and-post evolving feature spaces can be massively more complex 
than the previously explored linear models,
and must be gauged by a neural approximator 
that grows its capacity autonomously and adaptively.
%
%
%


\section{Preliminaries} \label{sec:preliminary}
 
We 
formulate the problem
in Section~\ref{subsec:problem},
present the challenges 
in Section~\ref{subsec:challenge},
and outline the key design ideas in Section~\ref{subsec:idea}.


\subsection{Problem Statement} \label{subsec:problem}
%

%

%
Let $\{ (\x{},y_t) \mid t=1,2,\ldots,T \}$ denote an input sequence,
where $\x{}$ is the data instance observed at the $t$-th round,
accompanied with a ground truth label $y_t \in \{ 1, 2, \ldots, C \}$.
It is worth noting that our online classification problem 
is formulated
in a multi-class regime with in total $C$ class options,
which excels our competitors~\cite{zhang2016online,hou2017learning,he2019online,he2021online} 
that focus on binary classification only.
%
%

\begin{figure}[t]
    \centering
    \includegraphics[width=.4\textwidth]{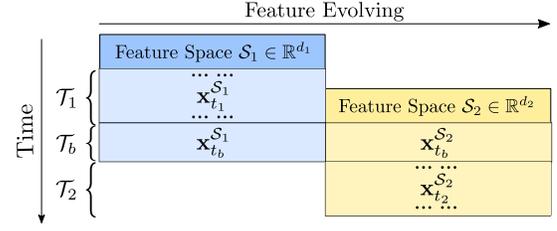}
    \caption{\label{fig:FESL}
        Illustration of doubly-streaming data. 
        %
        %
        Only in a very short timespan
        $\vert \Tset{b} \vert \ll \vert \Tset{1} \vert \textnormal{ or } \vert \Tset{2} \vert$,
        the samples are described by the 
        two feature spaces concurrently. 
        %
    }
    \vspace{-1em}
\end{figure}

In the context of doubly-streaming data,
we follow the pioneer~\cite{hou2017learning},
consider the set of features describing $\x{}$ 
to evolve with the following regularity,
illustrated in Figure~\ref{fig:FESL}.
Specifically,

\begin{itemize}
    \item In the span $t_1 \in \Tset{1} := \{ 1, \ldots, T_1 \} $, the classifier observes the instances described by the feature space $\S{1}$, \ie\ $\x{1} \in \S{1} \subseteq \Rset{d_1}$, each of which is a $d_1$-dimensional vector.
    

    \item In the span $t_b \in \Tset{b} := \{ T_1 + 1, \ldots, T_b \}$, the feature space evolves, and the classifier observes the two feature spaces $\S{1}$ and $\S{2}$ simultaneously, with each data instance being $\x{b} = [\x{b}^{\S{1}}, \x{b}^{\S{2}}]\T \in \S{1} \times \S{2} \subseteq \Rset{d_1 + d_2}$.
    

    \item In the span $t_2 \in \Tset{2} := \{ T_b + 1, \ldots, T_2 \}$, the old space $\S{1}$ opts out, and the classifier observes the evolved $\S{2}$ only. Each data instance is $\x{2} \in \S{2} \subseteq \Rset{d_2}$, a $d_2$-dimensional vector.
\end{itemize}

\noindent
Note, such feature space evolving from $\S{1}$ to $\S{2}$
can be easily generalized to infinitely more spaces
(\eg\ $\S{2}$ to $\S{3}$, then $\S{3}$ to $\S{4}$),
wherein all spaces can have disparate properties 
and semantic meanings 
and the mapping relationship between any two spaces 
can be arbitrarily complex.  
Such dynamism in the doubly-streaming data 
makes a prefix of learner capacity close to impossible.

%
At any time instant $t = \{ t_1, t_b, t_2\}$,
the learner $f_t$ observes $\x{}$ and makes a prediction
$\hat{y}_t = f_t (\x{})$. 
The true label $y_t$ is revealed thereafter,
and an instantaneous loss indicating the discrepancy between 
$y_t$ and $\hat{y}_t$ is suffered.
Based on the loss information,
the learner updates to $f_{t+1}$ using first-order~\cite{srebro2011universality,mcmahan2011follow,orabona2015generalized,fang2020online} or second-order~\cite{agarwal2006algorithms,hazan2007logarithmic,schraudolph2007stochastic,ye2018hones} oracles,
getting prepared for the next round.
Our goal is to find a sequence of classifiers $\{ f_1, \ldots, f_T \}$ 
that minimize the {\em empirical risk}~\cite{cesa2006prediction} 
over $T$ rounds:
$\min_{f_1, \ldots, f_T} \frac{1}{T}\sum_{t=1}^T \ell \big(y_t, f_t(\mathbf{x}_t) \big)$,
%
where $\ell(\argIn,\argIn)$ denotes the loss metric
and often is prescribed as convex in its argument 
such as square loss or logistic loss.

\subsection{Opportunities and Challenges}
\label{subsec:challenge}

A common practice to enable online learning 
with doubly-streaming data 
is to leverage the overlapping timespan $\Tset{b}$
to learn a \emph{reconstructive mapping}
$\phi:~\S{2} \mapsto \S{1}$,
such that 
once the features of $\S{1}$ are not observed during $\Tset{2}$,
their information can be reproduced,
allowing the learner to harvest the old information 
learned during the $\Tset{1}$ time period for better performance~\cite{hou2017learning,he2019online,he2021online,he2021onlineICDM}.

Let $f_{t} = \{ f_t^{\S{1}}, f_t^{\S{2}} \}$
denote the learner with 
$f_t^{\S{1}}$ and $f_t^{\S{2}}$ being the two classifiers 
corresponding to the $\S{1}$ and $\S{2}$ feature spaces, respectively.
During $\Tset{2}$, instead of predicting the observed instance 
as $f_t^{\S{2}}(\x{})$,
the learner exploits the unobserved information from $\S{1}$
to make prediction as:
$f_t(\vec{x}_t) = \lambda_1 \cdot f_t^{\S{1}}(\tilde{\vec{x}}_t) + \lambda_2 \cdot f_t^{\S{2}}(\x{})$,
with $\tilde{\vec{x}}_{t} = \phi(\x{}) \in \S{1}$
being the reconstructed data vector in the $\S{1}$ space.
With delicately tailored ensemble parameters $\lambda_1$ and $\lambda_2$,
this reconstruction-based learning method enjoys a provably 
better prediction performance than using the classifier $f_t^{\S{2}}$ only.


Unfortunately, this method 
does not scale up to cope with
real-world media data streams
because of two challenges as follows.


%


\subsubsection*{\bf Challenge \rom{1} -- Train Deep Models On-The-Fly}
\begin{figure}[!t]
	\centering
    \begin{subfigure}[t]{0.47\linewidth}
		
		\includegraphics[width=\textwidth]{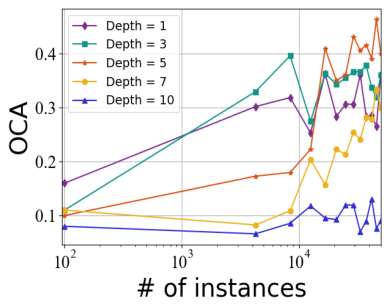}
		
	\end{subfigure}
	\begin{subfigure}[t]{0.47\linewidth}
		
		\includegraphics[width=\textwidth]{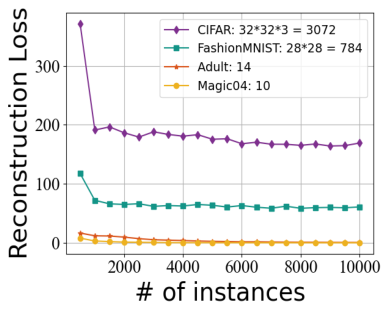}
		
	\end{subfigure}
    \caption{
        Two challenges underlie the \myAlg\ problem.
        {\em Left}: The deeper the learning model,
        the slower the convergence rate.
        {\em Right}: The higher the data dimensionality,
        the more inferior the feature relationship 
        captured by linear mappings.
    }
        \label{fig:challenge}
        \vspace{-1.5em}
    \end{figure}

The real-world media data carrying non-linear patterns
often request deep learners (\eg\ neural network models) 
for effective processing.
However, 
%
%
the large number of trainable parameters 
and complex model architectures 
tend to make deep learners data-hungry and converge slowly.
In an online learning context,
as each instance requiring immediate prediction
is presented only once,
the deep learners tend to {\em regret}~\cite{cesa2006prediction},
making substantial errors before converging to equilibria.
%
To verify this,
a simple example reduced from the CIFAR experiment
is illustrated in the left panel of Figure~\ref{fig:challenge},
where neural networks with various depths are trained 
in one-pass.

This example suggests that,
as the model depth goes deeper,
the learner suffers from a flatter convergence rate. 
Although such deep learners can end up with 
high online classification accuracy (OCA),
they constantly underperform shallower models 
before given sufficient instances,
thereby regretting largely.
Notably, a learner with an improperly ultra-deep 
architecture (\cf\ depth = 10) 
may even fail to converge in an online setting.
%
%
The reason can be possibly attributed to the 
diminishing feature reuse~\cite{huang2016deep,larsson2016fractalnet}
where the semantic meanings of raw inputs
tend to be \emph{washed out} by the layer-by-layer 
feedforward with massive randomly initialized parameters;
No expressive representations can be learned online.

%

\subsubsection*{\bf Challenge \rom{2} -- Learn Complex Reconstructive Mapping in Short Overlapping Timespans}
In practice, an overlapping phase $\Tset{b}$
in which the two feature spaces $\S{1}$ and $\S{2}$ coexist
is very short.
Revisit the smart manufacturing example,
where we can construct $\Tset{b}$
by pre-deploying a batch of new sensors 
before the old sensors expiring their lifespans --
a too long $\Tset{b}$ is economically not affordable.
This constraint 
blocks several seemingly plausible methods, \eg\
online transfer learning~\cite{zhao2014online,wu2017online},
domain adaptation~\cite{jain2011online,pirk2020online},
to work well, as they all require 
a sufficiently long overlapping phase to {\em align} 
the features pre and post evolution.

Prior studies~\cite{hou2017learning,he2019online,he2021online} 
have advocated deducing \emph{linear} functions 
to approximate the mapping relationship $\phi$
between the old and new features in a short $\Tset{b}$,
with the objective formulated as: 
$\min_{\phi} \sum_{t_b = T_1 + 1}^{T_b} \left \Vert \phi \big(\x{b}^{\S{2}} \big)  - \x{b}^{\S{1}} \right\Vert_2^2 
\textnormal{ where }
\phi \big( \argIn \big) = \vec{W}\T \argIn$.
%
Unfortunately, this linear reconstructive mapping $\phi$
cannot work for media data streams with 
nonlinear feature interplay.
An empirical evidence is presented in the right panel of 
Figure~\ref{fig:challenge},
in which we observe that,
the higher the data dimension,
the more complex the mapping relationship between 
two feature spaces,
and hence the larger the reconstruction loss 
that a linear mapping suffers.
%

%

\begin{figure*}[!t]
\centering
\includegraphics[width=.723\textwidth]{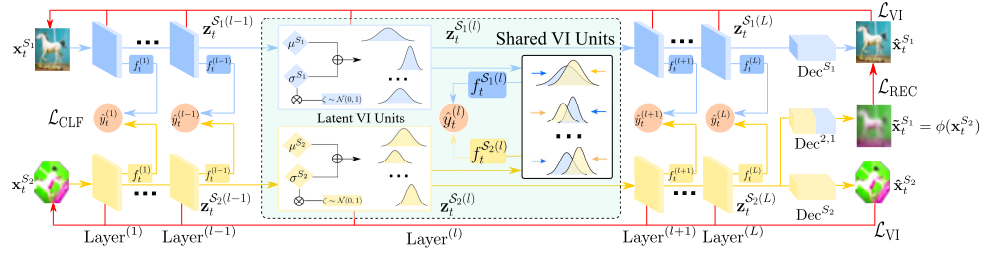}
\caption{An architectural illustration of our \myAlg\ computational network during the overlapping $\Tset{b}$ timespan. 
}
\label{fig:VAE}
\vspace{-1em}
\end{figure*}

\subsection{Our Thoughts} \label{subsec:idea}

To overcome the two challenges,
our key idea is to discover a set of {\em shared latent} features 
that summarize information from the pre-and-post evolving
feature spaces $\S{1}$ and $\S{2}$.
Compared with learning the 
mapping $\phi:~\S{2} \mapsto \S{1}$ directly 
in the short $\Tset{b}$,
our idea can exploit the long $\Tset{1}$ timespan
to learn a latent feature subspace
from $\S{1}$ independently at first,
and then align it with that from $\S{2}$ to expedite 
learning efficiency.
Specifically, we employ variational inference~\cite{blei2017variational} 
to model the underlying distribution of $\S{1}$ stream as:
\begin{equation} \label{eq:Gaussian}
    \small
	Q \left(\z{1}_{t_1} \mid 
    \x{1},~ t_1 = 1, \ldots, T_1  \right) = 
	\prod_{i=1}^{z} \mathcal{N} \left(\z{1}_i \mid \mu_i\TS{1}, (\sigma_i\TS{1})^2 \right),
\end{equation}
where a variational code $\vec{z}\TS{1} \in \Rset{z}$ is drawn
from a multivariate Gaussian that surrogates  
the data instances streaming from the
original feature space $\S{1}$~\cite{kingma2014autoencoding}.
%
Later in the overlapping $\Tset{b}$ phase,
a new variational code $\vec{z}\TS{2} \in \Rset{z}$
is extracted from the $\S{2}$ stream,
similar as \eqref{eq:Gaussian} and omitted for simplicity.
%
The two surrogate Gaussians that
approximate the $\S{1}$ and $\S{2}$ distributions 
(from which $\vec{z}\TS{1}$ and $\vec{z}\TS{2}$ were drawn)
are enforced to be identical,
such that they can be deemed as the shared latent subspace 
that connects the old and new feature spaces.
As such, we intermediately 
reconstruct the $\S{1}$ data representations 
from the shared surrogate statistics.

To make this process online,
we propose a neural architectural design 
which \emph{learns} the optimal model depth 
from data streams autonomously,
starting from shallow and gradually turning to deep 
if more complex variational feature mapping relationships 
are required to be approximated.
The more accurate this reconstructive mapping is approximated,
the better the learner can leverage
the old classifier trained on the $\S{1}$ stream,
and hence the higher the online classification accuracy
can be obtained by ensembling the old and new classifiers.
The details are presented 
in the next Section~\ref{sec:method}.

\section{Our Approach} 
\label{sec:method}

\par\smallskip\noindent
\noindent{\bf Overview.}
In a nutshell,
our proposed \myAlg\ approach can be conceptually framed 
in a learning objective as follows.
\begin{equation*}
	\min_{f_t, \phi} \sum_{\textnormal{HBP}} 
	\Big[ 
	\sum_{t_1, t_b} 
	\Big( \mathcal{L}_{\textnormal{VI}} (\phi)
	+ \mathcal{L}_{\textnormal{REC}} (\phi) \Big)
	+ \sum_{t_b,t_2} \mathcal{L}_{\textnormal{CLF}} (f_t, \phi)
	\Big].
\end{equation*}
In this section, we scrutinize this learning objective in sequence.
The variational inference loss 
$\mathcal{L}_{\textnormal{VI}}$ and the reconstruction loss
$\mathcal{L}_{\textnormal{REC}}$
together determine how 
the shared latent subspace is learned, 
presented in Section~\ref{subsec:VAE}.
The classification loss $\mathcal{L}_{\textnormal{CLF}}$
synopsizes how the old and new classifiers
are ensembled to expedite convergence
for better prediction performance in Section~\ref{subsec:ensemble}.
%
%
%
We end this section by elaborating how this minimization 
problem is realized by an {\em elastic} neural network model
that automatically adjusts its depth in an online, 
data-driven fashion
in Section~\ref{subsec:HBP}.



\subsection{Variational Latent Subspace Discovery}
\label{subsec:VAE}

%
To discover the latent subspace $\Z$,
we employ the Variational Auto-Encoder (VAE)~\cite{kingma2014autoencoding, blundell2015weight,galke2021lifelong}
to summarize the observed data instances 
into latent variational codes.
As illustrated in Figure~\ref{fig:VAE},
two independent VAEs are established, 
trained by minimizing the loss term:
\begin{equation} \label{eq:VIloss}
	\small
	\mathcal{L}_{\textnormal{VI}}^{ \{\S{1}, \S{2}\} } = 
	- \mathbb{E}_{Q(\vec{z}_t \mid \x{})} \big[\log \Prob(\x{} \mid \vec{z}_t) \big]
	+ \infdiv{Q(\vec{z}_{t}\mid \x{})}{\Prob(\vec{z}_{t})},
\end{equation}
where $t \in \Tset{1} \cup \Tset{b}$ 
and $t \in \Tset{b}$
for the VAEs on $\S{1}$ and $\S{2}$,
respectively.
%

\paragraph{\bf Intuition 1}
The physical meanings of minimizing 
\eqref{eq:VIloss} are as follows.
\romsm{1})
Minimizing the first term equates to maximizing the data generation quality,
namely, the likelihood that the original data observations
can be decoded from the extracted latent codes.
Let the tuple $(\enc,\dec)$ denote the encoder and the decoder networks in a VAE, 
the first term encourages $\x{} \approx \dec(\vec{z}_t)$
where $\vec{z}_t = \enc(\x{})$.
\romsm{2})
The second term gauges 
the Kullback-Leibler (KL)-divergence~\cite{blundell2015weight,kullback1951information}  
between the underlying posterior $Q(\vec{z}_{t}\mid \x{})$ 
and the latent marginal 
$\Prob(\vec{z}_{t}) = \mathcal{N}(\vec{0}, \vec{I})$.
With the posterior calculated by \eqref{eq:Gaussian},
for the extracted latent code $\vec{z}_{t}$,
we denote its $i$-th entry with $z_i$
and is drawn from a Gaussian with mean 
$\mu_i$ and variance $\sigma_i^2$.
To make the variational inference differentiable,
reparameterization is employed as 
$z_i = \mu_i + \sigma_i \cdot \zeta$ with 
$\zeta \sim \mathcal{N}(0,1)$ being normal noises.

A reconstruction loss is then imposed to regularize 
the two independently learned latent spaces,
from which a shared latent feature subspace is discovered
during the overlapping timespan $\Tset{b}$:

{\small
	\begin{equation} \label{eq:RECloss}
	\setlength{\abovedisplayskip}{-1em}
	\setlength{\belowdisplayskip}{-1.5em}
	\mathcal{L}_{\textnormal{REC}} = 
	\ell \left[ \x{b}^{\S{1}}, \dec^{2,1}(\z{2}_{t_b}) \right] +
	\infdiv{Q(\vec{z}_{t_b}\TS{1}\mid \x{b}\TS{1})}{Q(\vec{z}_{t_b}\TS{2}\mid \x{b}\TS{2})}.
\end{equation}}

\paragraph{\bf Intuition 2}
In the first term of \eqref{eq:RECloss}, 
a new decoder network $\dec^{2,1}(\argIn)$ which takes in 
the latent code from $\S{2}$ to 
reconstruct the data of $\S{1}$
approximates our desired 
reconstructive mapping $\phi$.
The second term gauges the KL-divergence between the posteriors
that were independently drawn from different variational distributions.
Minimizing this term encourages the different variational distributions 
--  the surrogate Gaussians 
to have similar probability densities,
as conceptually illustrated in the middle panel of Figure~\ref{fig:VAE}.
We note that this term is asymmetric, where the variational density
of $\S{2}$ is required to resemble that of $\S{1}$
but not the opposite.
This makes an intuitive sense as the variational distributions
of $\S{1}$ have been learned from $\Tset{1}$ over a long time horizon,
which is more likely to yield an accurate approximation
of the underlying data distribution than that from a much 
shorter $\Tset{b}$ only.

The two losses in Eqs.\,(\ref{eq:VIloss}) and (\ref{eq:RECloss}) 
together discover the shared latent feature subspace $\Z$.
In the subsequent $\Tset{2}$ timespan in which 
only the $\S{2}$ space can be observed, 
an arriving instance $\x{2}$ is embedded into $\Z$
by its corresponding VAE as 
$\vec{z}_{t_2} = \enc(\x{2})$,
from which a reconstructed data representation of the $\S{1}$ space
is decoded, \ie\ 
$\tilde{\vec{x}}_{t_2}^{\S{1}} := \phi(\x{2}) = \dec^{2,1}(\vec{z}_{t_2})$.
This VAE architecture lends to learn a
complex mapping relationship between $\S{1}$ and $\S{2}$,
hence 
better suits the high-dimensional media streams in the wild.

\subsection{Online Prediction with Ensembled Learners}
\label{subsec:ensemble}

Once the old features of $\S{1}$ vanish,
the learner $f_t$ is not likely to make accurate predictions on 
the arriving instances by relying on $\S{2}$ solely.
Let $f_{t} = \{ f_t^{\S{1}}, f_t^{\S{2}} \}$
denote the learner at the beginning of $\Tset{b}$
when $\S{2}$ just emerges.
As $\Tset{b}$ is short,
the $f_t^{\S{2}}$ part of the learner corresponding to the 
new features of $\S{2}$ have been trained with very few instances
hence is not likely to converge.
Relying on $f_t^{\S{2}}$ to predict the instances in $\Tset{2}$
would incur substantial regrets.

To aid, we leverage the old $f_t^{\S{1}}$ part 
that has been trained with a much larger number of instances 
during $\Tset{1}$. 
Thanks to the reconstructive mapping $\phi$
approximated by the VAEs in Section~\ref{subsec:VAE},
we can realize an online ensemble classification to yield 
accurate predictions when $f_t^{\S{2}}$ is not ready,
defined as follows.
{\small
	\begin{align} 
	\setlength{\abovedisplayskip}{-2.5em}
	\setlength{\belowdisplayskip}{-2em}
	&\mathcal{L}_{\textnormal{CLF}} := \ell (y_t, \hat{y}_t)
	= - \sum_{c=1}^C y_{t, c} \log(\hat{y}_{t, c}), 
	\quad \forall t \in \Tset{b} \cup \Tset{2},
	\label{eq:CEloss}\\
	&\hat{y}_t = p \cdot f_t^{\S{1}}(\xrec{t}) + 
	(1-p) \cdot f_t^{\S{2}}(\x{}), \quad \x{} \in \S{2},
	\label{eq:ensemble}
\end{align}}
where \eqref{eq:CEloss} employs cross-entropy~\cite{gao2017learning}
to gauge the multi-class learning loss,
with $y_{t, c}$ and $\hat{y}_{t, c}$ being the true
and predicted probability that $\x{}$
belongs to the $c$-th class, respectively.

\paragraph{\bf Intuition 3}
The idea behind \eqref{eq:ensemble}
is to let the ensemble coefficient $p \in (0,1)$ decide 
the impacts of the observed $\x{}$
and its reconstructed version $\xrec{t}$ in making predictions.
At the beginning of $\Tset{2}$ when the feature space just evolved,
the old classifier $f_t^{\S{1}}$ 
should be largely helpful with large $p$.
Over time, the value of $p$ decays because of two reasons
1) the new classifier $f_t^{\S{2}}$ becomes stronger and 
2) the old classifier $f_t^{\S{1}}$ can be less useful 
due to the distribution drift.
An updating strategy needs to be designed
to echo this intuitive process, where the new classifier
takes over gradually as the old classifier 
conveys less discriminative power.

In this work,
we update the ensemble coefficient
with exponential experts~\cite{cesa2006prediction},
where the empirical risks of using the old and new classifiers
to make independent predictions are accumulated as:
\begin{equation} \label{eq:S12Risk}
	\small
	R^{\TS{1}}_T = \sum_{t = T_1 + 1}^{T_2}  \ell \big(y_t, f_t^{\S{1}}(\xrec{t}) \big),
	~
	R^{\TS{2}}_T = \sum_{t = T_1 + 1}^{T_2}  \ell \big(y_t, f_t^{\S{2}}(\x{}) \big).
\end{equation}
The smaller the cumulative empirical risk is suffered,
the better the classifier is,
and hence the higher its corresponding coefficient is uplifted exponentially.
The updating rule is defined as
$p = e^{- \eta R^{\TS{1}}_T } / 
(e^{- \eta R^{\TS{1}}_T } + e^{- \eta R^{\TS{2}}_T })$,
%
where $\eta$ is a tuned parameter.

\subsection{Adaptive Model Depth Learning with HBP}
\label{subsec:HBP}

With the reconstructive mapping
and the ensemble prediction,
the information conveyed by the unobserved $\S{1}$
can be reaped to better the learning performance.
The remaining problem 
is how to realize the mapping and the classifiers with models of 
appropriate depths that are most likely to produce the optimal solutions.
Unfortunately, fixing such depths beforehand 
is impossible without prior knowledge of 
how the data streams evolve in the sample space
(\eg\ distribution drift that may require classifiers with various 
discriminant power to avoid overfitting)
and the feature space
(\eg\ a diversity of feature mapping relationships requires VAEs 
with disparate architectures).
As it is unrealistic to rely on human experts to provide 
such knowledge constantly over long timespans,
this problem boils down to the desire of a model architecture
that can \emph{learn} the best depth from data autonomously.

To this end,
we leverage the 
Hedge Backpropagation (HBP)~\cite{ijcai2018-369,he2021unsupervised}
mechanism to incorporate the model depth as a learnable semantic
that shall be determined in a data-driven manner 
through optimization.
Instead of evaluating the loss based on 
the output from the last network layer only
(as most deep learning models do),
the main idea of HBP is to evaluate 
the losses on {\em all} the intermediate hidden representations
yielded from the network layers from shallow to deep.
%
%
Specifically, given an overcomplete network with $L$ hidden layers in total,
the output of the $l$-th encoder layer of the VAE 
is recursively denoted as
$\vec{z}_t^{(l)} = \enc^{(l)} \big(\vec{z}_t^{(l-1)} \big),
\textnormal{ with }
\vec{z}_t^{(0)} = \x{}$,
%
where $t \in \Tset{1} \cup \Tset{b}  $ and $t \in \Tset{b} \cup \Tset{2}$ 
for the VAEs corresponds to $\S{1}$ and $\S{2}$,
respectively. The objective of HBP is defined as follows.
\begin{equation} \label{eq:HBP}
	\small
	\min_{\{ \alpha^{(l)} \}_{l=1}^L} \sum_{l=1}^L
	\alpha^{(l)} \bigg[
		\sum_{t_1, t_b} \Big( \mathcal{L}_{\textnormal{VI}}^{(l)} + \mathcal{L}_{\textnormal{REC}}^{(l)}  \Big) +
		\sum_{t_b, t_2} \mathcal{L}_{\textnormal{CLF}}^{(l)}
	\bigg],
\end{equation}
where the loss terms 
$\mathcal{L}_{\textnormal{VI}}^{(l)}$,
$\mathcal{L}_{\textnormal{REC}}^{(l)}$, and 
$\mathcal{L}_{\textnormal{CLF}}^{(l)}$
are evaluated on $\vec{z}_t^{(l)}$ at the $l$-th layer 
as shown in Figure~\ref{fig:VAE}.
In particular, 
1) Evaluated by 
$\mathcal{L}_{\textnormal{VI}}^{(l)}$ is 
how well the latent code $\vec{z}_t^{(l)}$ can summarize
the raw inputs with a surrogate Gaussian via using \eqref{eq:VIloss};
For instances of $\S{1}$, it is evaluated over $\Tset{1}$ and $\Tset{b}$ 
timespans, and for instances of $\S{2}$, 
it is evaluated over $\Tset{b}$ only.
2) Evaluated by 
$\mathcal{L}_{\textnormal{REC}}^{(l)}$ is 
how precisely the reconstructive mapping is learned 
so that the $\S{1}$ feature space can be reconstructed
from the data instances of $\S{2}$ via using \eqref{eq:RECloss};
It is only evaluated during the overlapping phase $\Tset{b}$ 
where $\S{1}$ and $\S{2}$ coexist.
3) Evaluated by 
$\mathcal{L}_{\textnormal{CLF}}^{(l)}$ is
how accurately the ensemble of both old and new classifiers
can make online predictions via using \eqref{eq:CEloss};
It is evaluated during $\Tset{b}$ and $\Tset{2}$ as the ensemble prediction 
is used only if the features of $\S{2}$ become observed.

\paragraph{\bf Intuition 4}
The crux of HBP lies in finding the equilibrium 
that minimizes the three loss terms in \eqref{eq:HBP}  
into a Pareto optimum.
To do this, we update the hedge weight $\alpha^{(l)}$
that determines the impact of the $l$-th layer 
in a boosting fashion~\cite{freund1997decision}:\
$\alpha^{(l)}_{t+1} \gets \textnormal{Norm} \big(\alpha^{(l)}_{t}
\beta^{\mathcal{L}_{\textnormal{VI+REC+CLF}}^{(l), t}} \big)$
,
%
%
where $\beta \in (0,1)$ is a discounting rate
and $\mathcal{L}_{\textnormal{VI+REC+CLF}}^{(l), t}$ accumulates 
the three losses in \eqref{eq:HBP} suffered at the $t$-th round.
Denoted by 
$\textnormal{Norm}(\argIn)$ is a normalization function
that reweighs each $\alpha^{(l)}$ by the sum of all $L$ layers,
ensuring $\alpha^{(l)}\in (0,1)$. 
The idea is straightforward:
the layer of which the output incurs large losses 
should be penalized and takes a discounted weight 
in the next round.
Otherwise, if a layer is in an \emph{optimal} depth,
it approaches the minimizer of \eqref{eq:HBP}
with the incurred losses very small,
such that the remaining layers (\ie\ those deeper than this hidden layer) 
cannot identify and learn meaningful gradient directions.
Their hedge weights would stay in small values.

\section{Experiments} \label{sec:experiments}

\begin{table}[!t]
	\centering
	\small
	\caption{Statistics of the 10 datasets.
		$| \S{1} |$ and $| \S{2} |$ are the dimensions of the old and new
		feature spaces, respectively. }
		\vspace{-1em}
	\begin{tabular}{c l|cccc}
		\toprule
		\midrule
		No. & Dataset & \# Samples   & $| \S{1} |$ & $| \S{2} |$ & \# Classes \\
		\midrule
		1   & magic04 & \num{36119}  & \num{10}    & \num{30}    & \num{2}    \\
		2   & adult   & \num{61559}  & \num{14}    & \num{30}    & \num{2}    \\
		\midrule
		3   & EN-FR   & \num{34758}  & \num{21531} & \num{24892} & \num{6}    \\
		4   & EN-IT   & \num{34758}  & \num{21531} & \num{15506} & \num{6}    \\
		5   & EN-SP   & \num{34758}  & \num{21531} & \num{11547} & \num{6}    \\
		6   & FR-IT   & \num{49648}  & \num{24893} & \num{15503} & \num{6}    \\
		7   & FR-SP   & \num{49648}  & \num{24893} & \num{11547} & \num{6}    \\
		\midrule
		8   & CIFAR   & \num{95000}  & \num{3072}  & \num{3072}  & \num{10}   \\
		9   & Fashion & \num{114000} & \num{784}   & \num{784}   & \num{10}   \\
		10  & SVHN    & \num{139257} & \num{3072}  & \num{3072}  & \num{10}   \\
		\bottomrule
	\end{tabular}
	\label{tab:dataset}
	\vspace{-2em}
\end{table}

Empirical results are presented
to verify the viability and effectiveness of our \myAlg\ approach.
We elaborate the experimental setups 
in Section~\ref{subsec:dataset}
and extrapolate the results and findings in Section~\ref{subsec:results}.

\subsection{Evaluation Setup}
\label{subsec:dataset}


\subsubsection{\bf Dataset Preparation}
We benchmark our \myAlg\ approach on 10 real-world datasets 
covering three domains to verify its versatility.
%
Statistics of the studied datasets are summarized in Table~\ref{tab:dataset}.
%

{\em \bfseries $\bullet$ UCI Data Science (No. 1-2):}
The two datasets have one feature space $\S{1}$ at first,
and we artificially create a new feature space
$\S{2} = \textnormal{sigmoid}(\vec{W}\T \S{1})$ with a random Gaussian $\vec{W}$ and a nonlinear sigmoid function.
The two feature spaces are concatenated as the shape in Figure~\ref{fig:FESL} to simulate the doubly streaming data.

{\em \bfseries $\bullet$ Multilingual Text Categorization (No. 3-7):}
A set of documents are described by four languages including English (EN), French (FR),
Italian (IT), and Spanish (SP).
By treating each document as a bag of words (features),
the vocabulary of each language can be deemed as a feature space.
At each time, a document is presented and our model aims to classify it into one of the six categories.
%
To simulate doubly-streaming,
the language describing the documents shifts over time, \eg\ EN-FR,
where the model learned to classify English documents is soon presented with French documents
after a short overlapping $\Tset{b}$ timespan,
requiring to approximate the translation relationship
between languages.
To exacerbate the non-linearity of the mapping between two languages,
we apply the sigmoid function on the $\S{2}$ feature space.
%

{\em \bfseries $\bullet$ Online Image Classification  (No. 8-10):}
Images are typical media data of high dimensionality and low information density.
To simulate doubly-streaming data,
we follow the preprocessing steps suggested by~\cite{oml,he2021unsupervised} to create an evolved space
by transforming the original images with
various spectral-mapping, shearing, rescaling, and rotating.
Images are presented one at a step,
and the model needs to learn the complex pixel transformation online.
%

\subsubsection{\bf Compared Methods}
Three state-of-the-art competitors
tailored for processing double-streaming data
are employed for comparative study,
with their main ideas presented as follows.

{\bfseries $\bullet$ FOBOS}~\cite{duchi2009efficient}
is a canonic online learning baseline
that operates over first-order oracles with a projected
subgradient that encourages sparse solutions.
%
To make it work for doubly-streaming data,
zeros are padded to the new features and vanished old features.

{\bfseries $\bullet$ OLSF}~\cite{zhang2015towards}
is the first study to tackle an incremental feature space,
where new features constantly emerging
are carried in all subsequent data instances.
OLSF updates the online learners in a passive-aggressive fashion,
where the learning coefficients of old features
are re-weighed to new features only if these new features convey
significant information that changes the decision boundary.
%

\begin{table*}[!t]
	\centering
	\small
	\setlength{\tabcolsep}{15pt}
	\caption{\label{tab:errors}
		Comparative results of averaged cumulative regret
		(ACR $\pm$ mean variance) benchmarked on 10 datasets,
		where the lower the value,
		the better the method performs.
		The best results are bold.
		The bullet $\bullet$ indicates that our \myAlg\ approach
		outperforms the competitors with a statistical significance
		supported by the
		\textit{paired t-tests} at $95\%$ confidence level.
	}
	\vspace{-.5em}
	\begin{tabular}{l|c|c|c|c|c|c}
		\toprule
		\midrule
		Dataset & FOBOS                                     & OLSF                                      & FESL                                      & OLD-Linear & OLD-FD & \myAlg\ \\
		\midrule

		magic04 & $.119\pm.022\bullet$                      & $.335\pm.021\bullet$                      & $.110\pm.016\bullet$
		        & $.075\pm.018$\textcolor{white}{$\bullet$} & $.076\pm.021$\textcolor{white}{$\bullet$} & $\mathbf{.052\pm.017}$
		\\

		adult   & $.076\pm.064$\textcolor{white}{$\bullet$} & $.225\pm.019\bullet$                      & $.067\pm.044$\textcolor{white}{$\bullet$}
		        & $.055\pm.017$\textcolor{white}{$\bullet$} & $.068\pm.018$\textcolor{white}{$\bullet$} & $\mathbf{.049\pm.019}$
		\\

		\midrule

		EN-FR   & $.326\pm.064\bullet$                      & $.324\pm.018\bullet$                      & $.345\pm.044\bullet$
		        & $.168\pm.030\bullet$                      & $.137\pm.030\bullet$                      & $\mathbf{.068\pm.025} $
		\\

		EN-IT   & $.318\pm.060\bullet$                      & $.314\pm.019\bullet$                      & $.337\pm.040\bullet$
		        & $.197\pm.028\bullet$                      & $.143\pm.033\bullet$                      & $ \mathbf{.083\pm.024}$
		\\

		EN-SP   & $.302\pm.060\bullet$                      & $.322\pm.021\bullet$                      & $.335\pm.037\bullet$
		        & $.197\pm.036\bullet$                      & $.136\pm.027\bullet$                      & $\mathbf{.077\pm.024}$
		\\

		FR-IT   & $.278\pm.047\bullet$                      & $.301\pm.013\bullet$                      & $.314\pm.037\bullet$
		        & $.195\pm.031\bullet$                      & $.147\pm.030\bullet$                      & $\mathbf{.084\pm.026}$

		\\
		FR-SP   & $.272\pm.046\bullet$                      & $.310\pm.014\bullet$                      & $.336\pm.040\bullet$
		        & $.201\pm.029\bullet$                      & $.155\pm.026$\textcolor{white}{$\bullet$} & $\mathbf{.102\pm.027}$

		\\
		\midrule
		CIFAR   & $.468\pm.017\bullet$                      & $.504\pm.014\bullet$                      & $.463\pm.013\bullet$
		        & $.166\pm.032$\textcolor{white}{$\bullet$} & $.232\pm.038\bullet$                      & $\mathbf{.150\pm.030}$
		\\
		Fashion & $.305\pm.033\bullet$                      & $.294\pm.016\bullet$                      & $.247\pm.023\bullet$
		        & $.160\pm.033\bullet$                      & $.123\pm.019\bullet$                      & $\mathbf{.056\pm.015}$
		\\
		SVHN    & $.808\pm.011\bullet$                      & $.604\pm.014\bullet$                      & $.806\pm.011\bullet$
		        & $.144\pm.038\bullet$                      & $.120\pm.025$\textcolor{white}{$\bullet$} & $\mathbf{.089\pm.018 }$

		\\
		\midrule
		w/t/l   & 9/1/0                                     & 10/0/0                                    & 9/1/0                                     & 7/3/0      & 6/4/0  & ---      \\
		\bottomrule
	\end{tabular}

\end{table*}

{\bfseries $\bullet$ FESL}~\cite{hou2017learning}
is the pioneer work to deal with doubly-streaming data,
%
which nevertheless employed linear functions to learn classifiers
and to approximate a mapping relationship between feature spaces.
%
A comparison with FESL rationalizes our design of adaptive deep learner
and variational feature mapping approximator.

\subsubsection{\bf Ablation Variants}
For the ablation study, two variants of our \myAlg\ approach
are proposed, named OLD-Linear and OLD-FD.
They differ from our original \myAlg\ design by:
1) {\bf OLD-Linear} employed linear mapping to approximate
the feature mapping relationship
and
2) {\bf OLD-FD} trains a deep neural network with a fixed depth.
%
We craft the two variants to necessitate the designs
of a non-linear, VI-based feature mapping approximator and
the HBP that allows model depth to be learned from data autonomously.

\subsubsection{\bf Evaluation Metric}
As the traditional classification accuracy is ill-conditioned
in online learning, we employ the Online Classification Accuracy
({\bf OCA}) and Averaged Cumulative Regret ({\bf ACR})
to measure the performance.
Specifically, they are defined:
{\small
	\begin{align*}
	\setlength{\abovedisplayskip}{.1pt}
	\setlength{\belowdisplayskip}{.1pt}
	\textnormal{OCA}(f_t) & = 1 - \frac{1}{B} \sum_{i = t-B}^t
	\llbracket y_i \neq f_t (\vec{x}_i) \rrbracket, \quad
	T =\vert \Tset{1} \cup \Tset{b} \cup \Tset{2} \vert                                                     
	\\
	\text{ACR}            & = \frac{1}{T} \sum_{t=1}^{T} \Big[ \max_{f*} \textnormal{OCA}(f^*)
		- \textnormal{OCA}(f_t) \Big].
\end{align*}}
Intuitively, OCA dynamically measures
the accuracy of a classifier $f_t$ the $t$-th round,
evaluated at the {\em most recent} $B$ instances.
%
%
%
ACR evaluates how large the online learner regrets
comparing to a hindsight optimum $f^*$
by accumulating the OCA differences between $f_t$ and $f^*$
over $T$ rounds.
%
%
%
The smaller the value of ACR,
and the better the online classification was performed.

\subsection{Results and Findings}
\label{subsec:results}

\begin{figure*}[!t]
	\centering

	\begin{subfigure}[t]{0.1965\linewidth}
		\includegraphics[width=\textwidth]{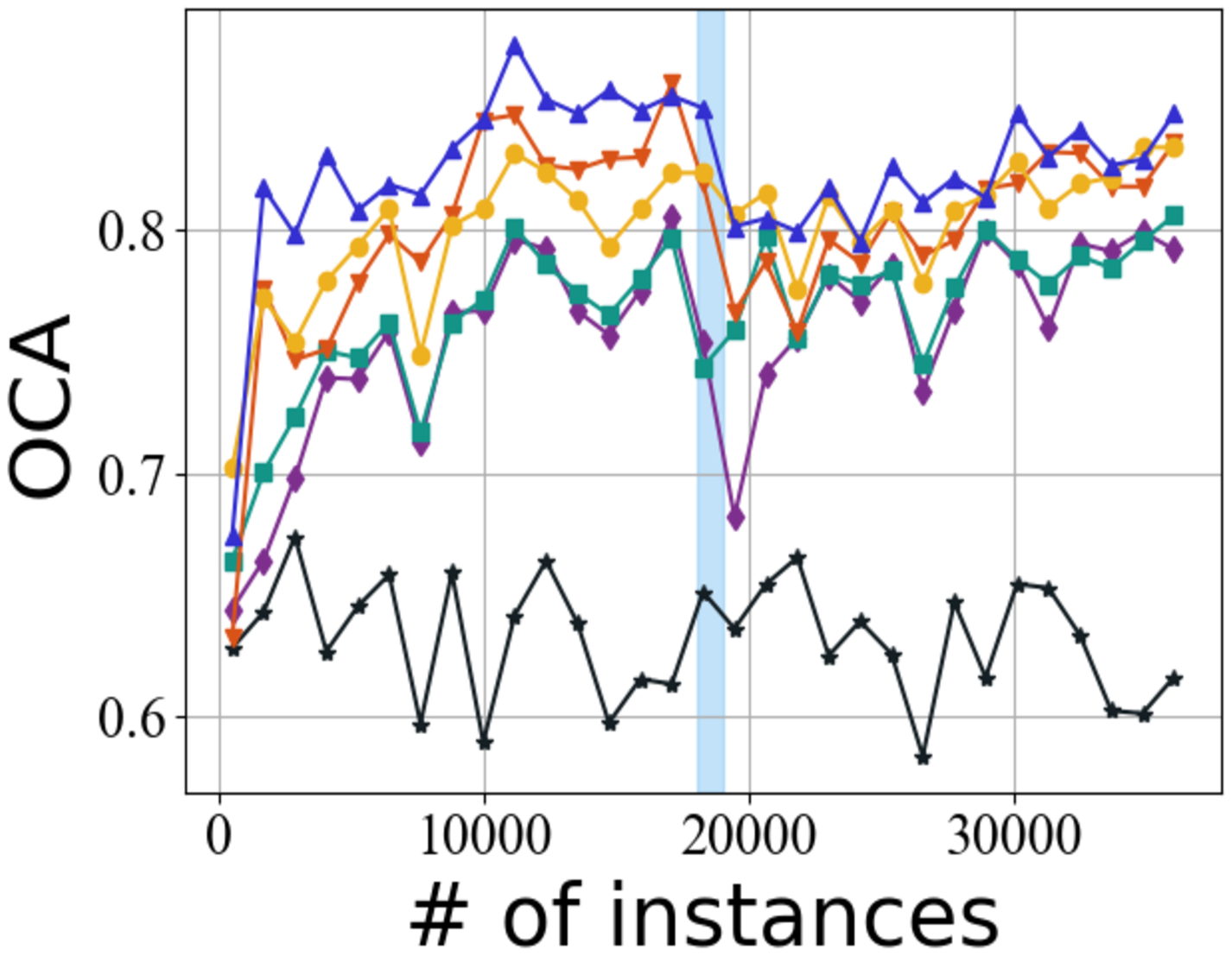}
		\caption{magic04}
		\label{fig:magic04}
	\end{subfigure}
	\begin{subfigure}[t]{0.1965\linewidth}
		\includegraphics[width=\textwidth]{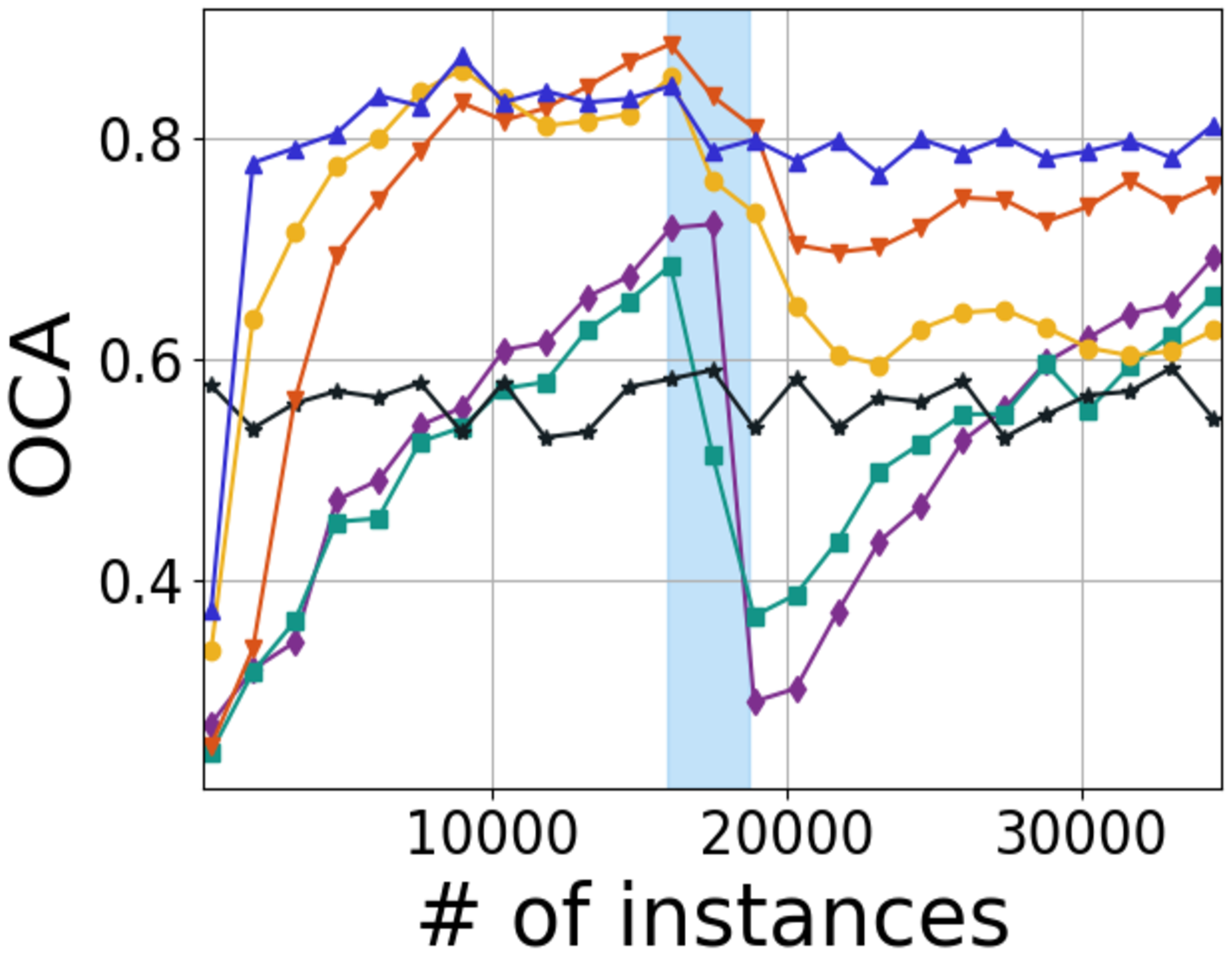}
		\caption{Reuter-EN-FR}
		\label{fig:reuterenfr}
	\end{subfigure}
	\begin{subfigure}[t]{0.1965\linewidth}
		\includegraphics[width=\textwidth]{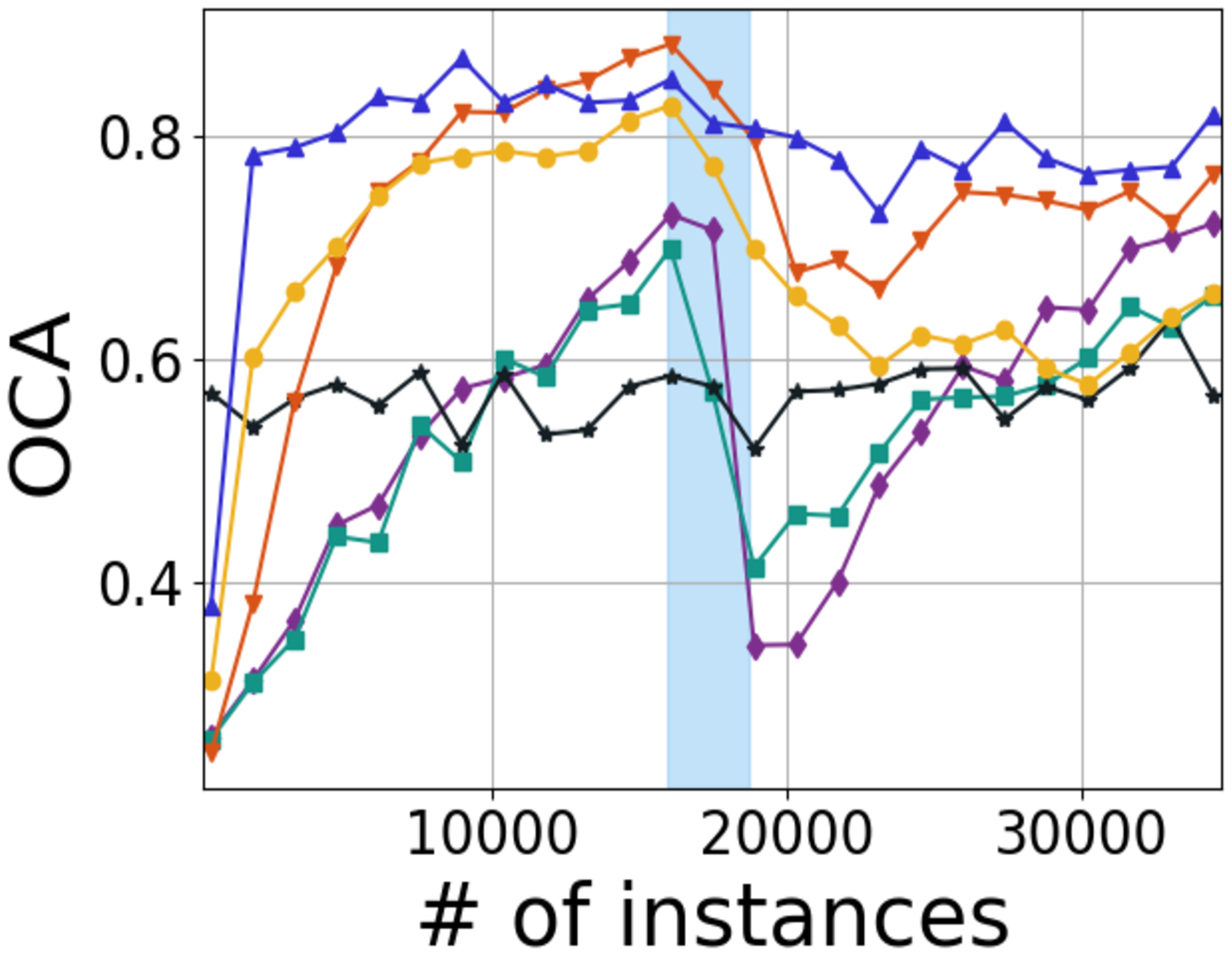}
		\caption{Reuter-EN-IT
		}
		\label{fig:reuterenit}
	\end{subfigure}
	\begin{subfigure}[t]{0.1965\linewidth}
		\includegraphics[width=\textwidth]{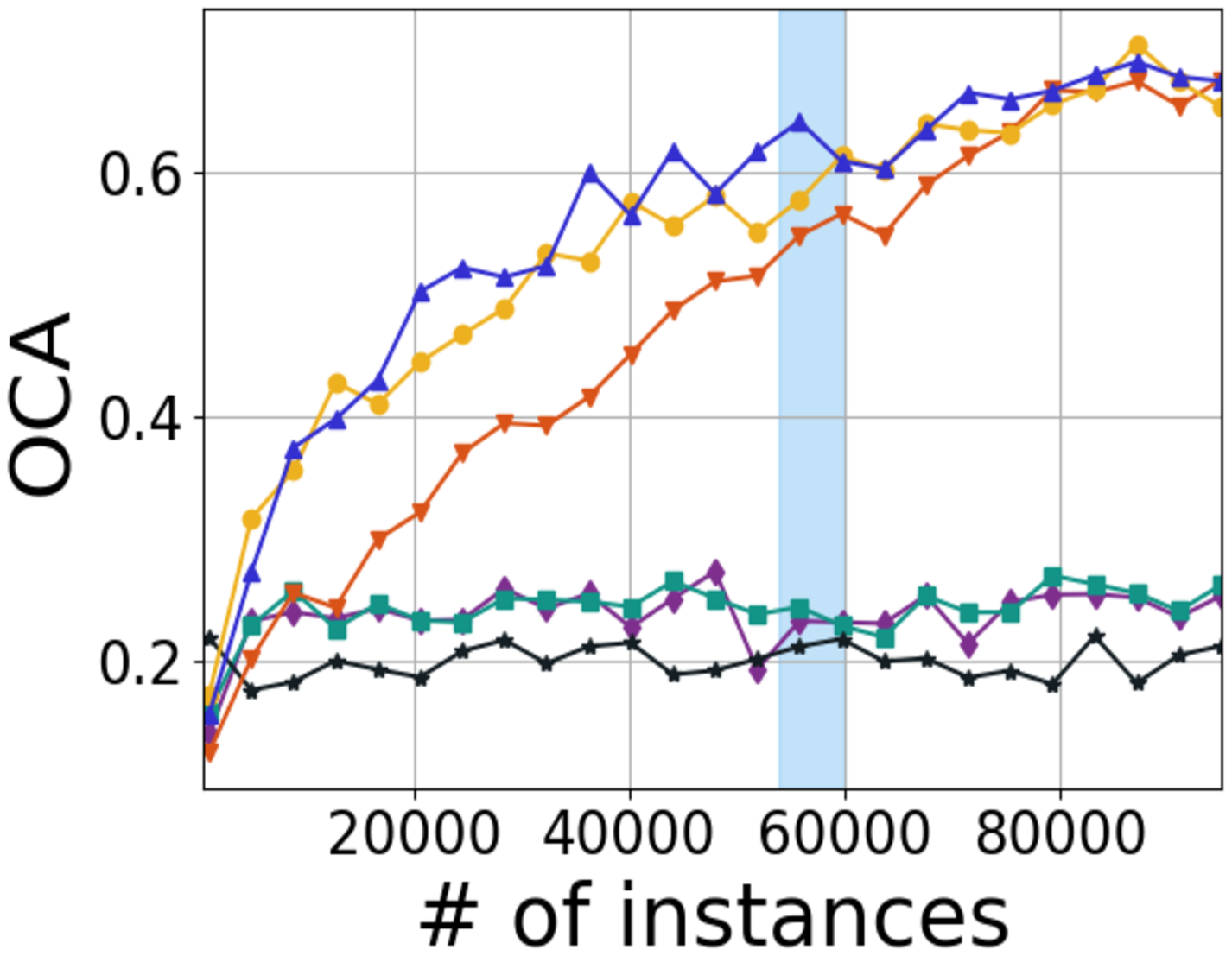}
		\caption{CIFAR
		}
		\label{fig:cifar}
	\end{subfigure}
	\begin{subfigure}[t]{0.1965\linewidth}
		\includegraphics[width=\textwidth]{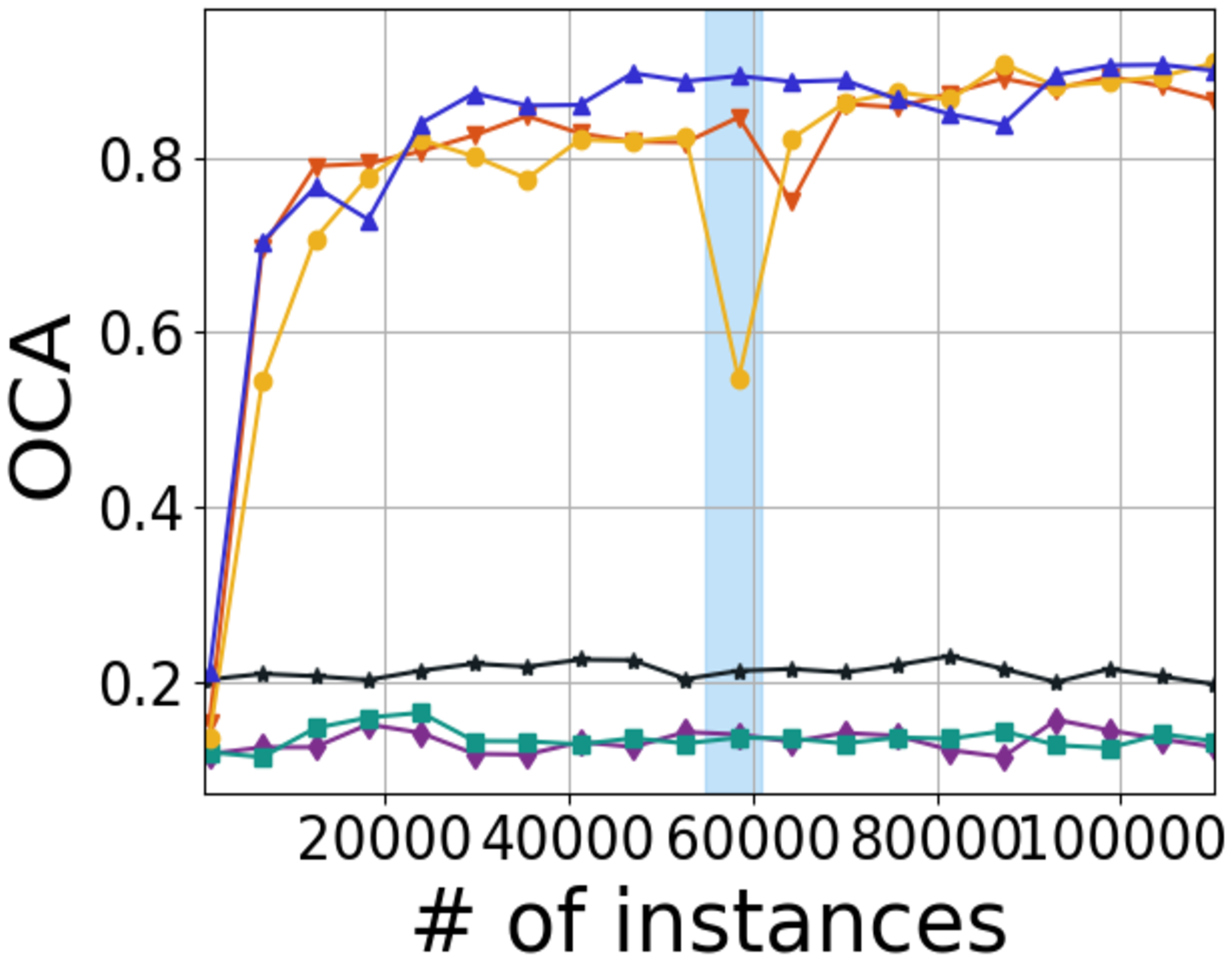}
		\caption{SVHN
		}
		\label{fig:svhn}
	\end{subfigure}

\vspace{.2em}

	\begin{subfigure}{0.7\linewidth}
		\includegraphics[width=\textwidth]{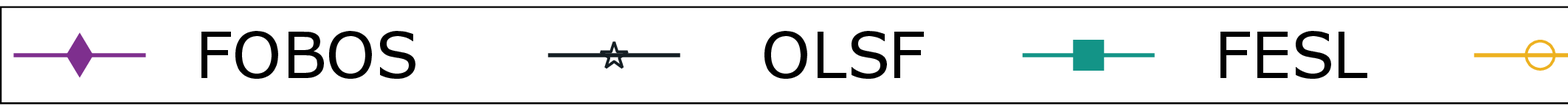}
		\label{fig:legend}
	\end{subfigure}
	\vspace{-2em}
\caption{
	The trends of OCA of six methods on five datasets
	in the doubly-streaming setting.
	The blue-shadowed areas indicate the overlapping $\Tset{b}$
	timespans.
	Due to the space limitation, 
	complete results are deferred to the supplementary file.
}
	\label{fig:results}
\end{figure*}

We present the experimental results in Table~\ref{tab:errors} and Figure~\ref{fig:results},
aiming to answer three
research questions ({\bf Q1} -- {\bf Q3}) as follows.

\begin{description}
	\item[Q1.] {\em How does our \myAlg\ approach compare to the state-of-the-arts?}
\end{description}

From the comparative results
presented in Table~\ref{tab:errors},
we make three observations as follows.
	{\em First}, our \myAlg\ achieves the best ACR performance.
%
This result rationalizes our proposal of learning deep learners
with complex feature relationships,
as the competitors mainly relying on linear models manifest
inferior performances.
	{\em Second},
our \myAlg\ outperforms FOBOS by \blue{69}\% on average.
In addition, FOBOS suffers the largest performance drop
in terms of OCA when the old features become unobserved,
as shown in Figures~\ref{fig:magic04},~\ref{fig:reuterenfr},and~\ref{fig:reuterenit}.
This is because that
FOBOS does not correlate the old and new feature spaces thus
can be equated to initializing a new learner for the newly emerged features.
Our approach excels as we learned the feature correlation
to boost the learning performance on the new features,
and then enjoys a much smoother learning curve as soon as the feature space evolves.

{\em Third},
compared to OLSF, our approach wins by \blue{77}\% on average.
The reason can be attributed to that OLSF is tailored for
dealing with an incrementally increasing feature space only,
and does not possess the mechanism to handle the fading away features.
The learned knowledge of the old feature space is hence wasted.
Our approach aids the situation
by learning a reconstructive mapping
between the two feature spaces,
letting the learner enjoy the information
conveyed by the old and unobservable features,
thereby attaining better ACR and sharper OCA curves along the time horizon.

\begin{description}
	\item[Q2.] {\em How helpful is the deep learner enabled by the VI mapping?}
\end{description}

The comparison among FOBOS, FESL, our \myAlg\ approach
and its OLD-Linear variant
amounts to the answer.
{\em First},
our \myAlg\ outperforms FESL and OLD-Linear
by ratios of  \blue{69}\% and  \blue{44}\% on average, respectively.
This performance gap indicates
the non-linear mapping relationship between feature spaces
must be respected,
as FESL and OLD-Linear both employed linear functions
to approximate the reconstructive mapping.
{\em Second}, more significant OCA drops
are observed from OLD-Linear in Figures~\ref{fig:reuterenfr},~\ref{fig:reuterenit},and~\ref{fig:svhn}.
This result suggests that
the low-dimensional latent space resulted
from the variational encoding
does not suffice to simplify the complex
feature reconstruction relationships
to an extent that they can be approximated by
linear functions.

{\em Third},
we observe that FESL may even underperformed FOBOS
in terms of ACR, despite that
FESL suffers a smaller performance drop of OCA overtime.
This observation advocates that
FESL learned the feature relationship at a certain level,
but the linearity of the mapping function does suffice
to fully capture the complex feature interactions,
such that
the linear reconstruction of old features is helpful
at the beginning of $\Tset{2}$ (smaller OCA drop)
but soon becomes less useful overtime
(slower learning rate),
and eventually becomes \emph{noises} which negatively
affect the prediction accuracy,
ending up with inferiority to FOBOS.
In other words, it is better to initialize a new learner
than trying to reconstruct old features inaccurately
with an insufficiently capable linear mapping.

\begin{description}
	\item[Q3.] {\em In which cases does an adaptive learning capacity  excel?}
\end{description}

A comparison between our \myAlg\ with the OLD-FD variant
answers this question.
We observe that 1) \myAlg\ excels and
significantly outperforms OLD-FD in six settings
2) \myAlg\ converges faster with steeper OCA curves in all settings.
These two observations validate the \emph{tightness}
of HBP
in the sense that, although
OLD-FD may end up with higher OCA with increasingly
more arriving data instances
(\eg\ Figures~\ref{fig:magic04} and~\ref{fig:cifar}),
its slower convergence rate incurs larger online prediction errors
before the network parameters are readily trained.
This necessitates the usage of HBP to expedite the online learning efficiency.

In addition, from Figures~\ref{fig:cifar} and~\ref{fig:svhn},
we observe that OLD-FD learns slower as the learning task becomes more difficult.
(The objects in CIFAR impose more complex visual concepts 
than the street-view numbers in SVHN, 
where the hindsight optimal OCAs in CIFAR and SVHN 
are \blue{72.7}\% and \blue{93.3}\%, respectively).
Our \myAlg\ is invariant to the inherent complexity of the datasets 
and manifests a fast online learning rate.
This finding advocates the adaptive model capacity
of our \myAlg\ is generalizable to more learning tasks, 
without requiring prior knowledge of the underlying distribution or 
learning complexity of the doubly-streaming data of interest.

\section{Conclusion}

This paper proposed a new online learning paradigm,
named \myAlg, which enables a deep learner
to make on-the-fly decisions on data streams 
with a constantly evolving feature space.
The key idea is to establish a mapping relationship 
between the old and new features,
such that once the old features vanish,
they are reconstructed from the new features,
allowing the learner to harvest both old and new feature information 
to make accurate online predictions via ensembling.
To realize this idea, the crux lies in the harmonization of model
onlineness and expressiveness.
To respect the high dimensionality and complex feature interplay
in the real-world data streams,
our \myAlg\ approach discovered a shared latent subspace 
using variational approximation, 
which can encode arbitrarily expressive mapping functions for feature reconstruction.
Meanwhile, as the real-time nature of data streams biases 
shallow models,
our approach enjoyed an optimal depth {\em learned} from data,
starting from shallow and gradually becoming deep if more complex patterns are
required to be captured in an online fashion.
%
%
Comparative studies evidenced the viability of our approach 
and its superiority over the state-of-the-art competitors.

%

%


\bibliographystyle{ACM-Reference-Format}
\bibliography{bibfile}
\clearpage




\end{document}